\def\year{2020}\relax
\documentclass[letterpaper]{article} 
\usepackage{aaai20}  
\usepackage{times}  
\usepackage{helvet} 
\usepackage{courier}  
\usepackage[hyphens]{url}  
\usepackage{graphicx} 
\usepackage{graphicx}  
\usepackage{makecell}
\usepackage{multirow}
\usepackage{amsmath}
\usepackage{amssymb}
\usepackage{enumitem}
\usepackage{color}
\newcolumntype{L}[1]{>{\raggedright\let\newline\\\arraybackslash\hspace{0pt}}m{#1}}
\title{Relevance-Promoting Language Model for Short-Text Conversation\thanks{The work described in this paper is substantially supported by a grant from the Research Grant Council of the Hong Kong Special Administrative Region, China (Project Code: 14204418). It was mainly done when Xin Li was an intern at Tencent AI Lab.}}
\author{
Xin Li,\textsuperscript{\rm 1}
Piji Li,\textsuperscript{\rm 2}
Wei Bi,\textsuperscript{\rm 2}
Xiaojiang Liu,\textsuperscript{\rm 2}
Wai Lam\textsuperscript{\rm 1}\\
\textsuperscript{\rm 1}Department of Systems Engineering and Engineering Management,\\ The Chinese University of Hong Kong\\
\textsuperscript{\rm 2}Tencent AI Lab, Shenzhen, China \\
\{lixin, wlam\}@se.cuhk.edu.hk, \{pijili, victoriabi, kieranliu\}@tencent.com
}
 
\begin{document}

\maketitle

\begin{abstract}
Despite the effectiveness of sequence-to-sequence framework on the task of Short-Text Conversation (STC), the issue of under-exploitation of training data (i.e., the supervision signals from query text is \textit{ignored}) still remains unresolved. Also, the adopted \textit{maximization}-based decoding strategies, inclined to generating the generic responses or responses with repetition, are unsuited to the STC task. In this paper, we propose to formulate the STC task as a language modeling problem and tailor-make a training strategy to adapt a language model for response generation. To enhance generation performance, we design a relevance-promoting transformer language model, which performs additional supervised source attention after the self-attention to increase the importance of informative query tokens in calculating the token-level representation. The model further refines the query representation with relevance clues inferred from its multiple references during training. In testing, we adopt a \textit{randomization-over-maximization} strategy to reduce the generation of generic responses. Experimental results on a large Chinese STC dataset demonstrate the superiority of the proposed model on relevance metrics and diversity metrics.\footnote{Code available at https://ai.tencent.com/ailab/nlp/dialogue/.}
\end{abstract}

\section{Introduction}
Short Text Conversation (STC)~\cite{shang-etal-2015-neural}, also known as single-turn chit-chat conversation, is a popular research topic in the field of natural language processing. It is usually formulated as a sequence translation problem~\cite{ritter-etal-2011-data,shang-etal-2015-neural} and the sequence-to-sequence encoder-decoder (\textsc{Seq2Seq}) framework~\cite{cho-etal-2014-learning,sutskever2014sequence,bahdanau2015neural} is applied for solving this problem. The decoder generates the responses token-by-token, conditioned on the compressed query representations from the encoder. Following this paradigm, many attempts have been conducted to refine the quality of the generated responses~\cite{li-etal-2016-diversity,xing2017topic,du-etal-2018-variational,tian-etal-2019-learning}.

Despite the effectiveness of these efforts, some intrinsic issues of \textsc{Seq2Seq}-based models still hinder further improvement of generation performance. Under the \textsc{Seq2Seq} formulation, the auto-regressive decoder is only trained on the gold-standard response text while the query text is ignored, leading to under-exploitation of the training data. Besides, the maximization-based decoding strategies adopted in existing models, such as beam search and greedy search, restrict the search space to the most frequent phrases and thus they have the tendency to generate the generic responses or repetitive responses with unnaturally high likelihood, degrading the conversational experience. 

GPT-2~\cite{radford2019language}, a recently proposed Transformer-based language model, provides an alternative solution for language generation. One advantage of GPT-2 is that the transformer language model can not only capture the context of arbitrary length but also make full use of the textual supervision signals because the generator is actually the language model itself. Moreover, GPT-2 adopts top-\textit{k} sampling~\cite{fan-etal-2018-hierarchical} to diversify the generated texts while preserving the relevance. Obviously, these characteristics are attractive and meaningful for solving the STC task, whose aim is to generate informative and diverse human-like responses given the user queries.  

However, due to the essence of language modeling, directly applying GPT-2 on the STC task, a conditional language generation task, may be insufficient because the language model is unable to discriminate the source (query) sentence and the target (response) sentence. The original experimental results of GPT-2 on the abstractive summarization task~\cite{nallapati-etal-2016-abstractive} also verify this claim. Another potential issue of adapting language model for the STC task comes from \textbf{\textit{recency bias}}~\cite{khandelwal-etal-2018-sharp} and \textbf{\textit{explanation-away}} effects~\cite{yu2017neural,holtzman2019curious}, where the language model has the tendency to rely overly on the immediate context and explain away from the long-term context\footnote{Long-term context in language model is roughly equivalent to the source information in \textsc{Seq2Seq} framework.}, yielding fluent but topically irrelevant responses.

\begin{figure}
    \centering
    \includegraphics[width=1\columnwidth]{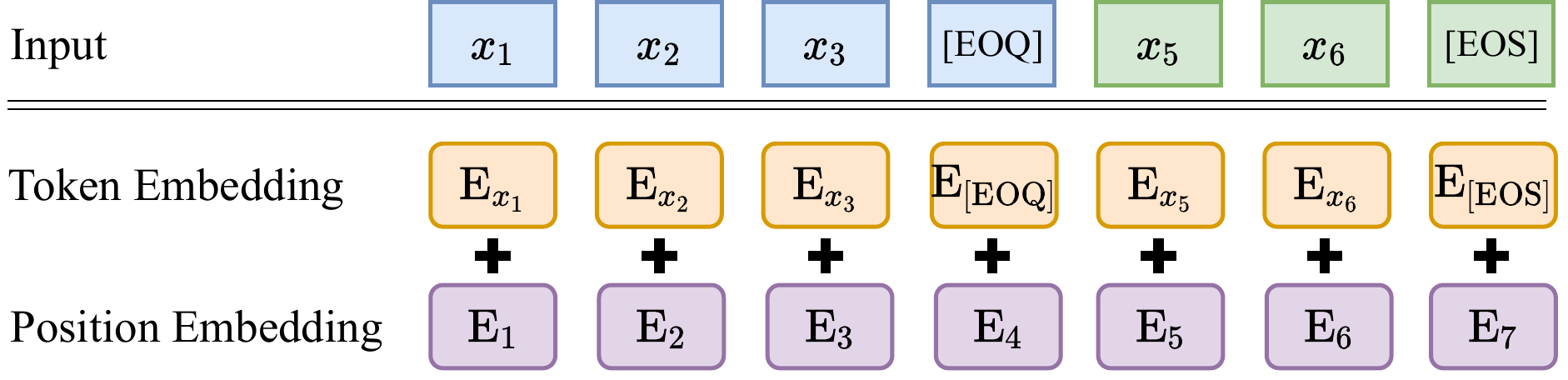}
    \caption{Representations of the example input with $n=7$ and $m=4$.}
    \label{fig:example_input}
\end{figure}

With the motivation of inheriting the merits of transformer language model while alleviating the potential issues under the language model formulation, we carefully design a training strategy to adapt the auto-regressive transformer-based language model\footnote{Without explicit specification, the language model in our paper refers to the ``auto-regressive'' language model, which is different from those ``auto-encoding'' language models~\cite{devlin-etal-2019-bert,dong2019unified}.} for the conditional response generation. First of all, it is observed that the dialog conversation is actually a process of text continuation, in other words, giving the response right after the query. Based on this observation, we can regard the STC task as a language modeling problem on the concatenated sequence of query and response. To discriminate the generation of query tokens and that of response tokens, we inject a special token between query and response, acting as the trigger of response generation. With this formulation, the language model based training objective can make use of the textual data from query, alleviating the under-exploitation issue mentioned above. 

Since the transformer-based language model tends to focus on the short-term context and ignore the long-term context, namely, the \textbf{\textit{explanation away}} issue, we propose to empower the self-attention with encoder-decoder attention, which enforces the model to pay additional attention to the query, especially the query tokens of user interest, and guides the model to rely on informative query tokens to make good predictions. It is also observed that some response tokens not mentioned in the query are still closely related to the discussed topic in the conversation. In order to exploit such kind of relevance clues hidden behind the responses, we propose a topic inference component to learn a compact source (query) representation encoding the information relevant to the query and feed the query representation into each generation step, encouraging the language model to consider the generation of the topic words potentially related to the query.  

As with the decoding strategy, different from the existing STC models, we propose to decode with \textbf{\textit{randomization-over-maximization}} method, namely, the top-\textit{k} sampling, from the transformer language model to generate the relevant response with high originality.

In summary, our contributions are as follows:

\indent $\bullet$ We tailor-make a training strategy to adapt the transformer-based language model for the Short Text Conversation (STC) task. \\
\indent $\bullet$ We propose two components, namely, Supervised Source Attention (SSA) component and Topic Inference (TI) component to promote the relevance modeling in the language model based response generator. \\
\indent $\bullet$ To the best of our knowledge, we are the first to introduce top-\textit{k} sampling, a \textbf{\textit{randomization-over-maximization}} strategy, for diverse response generation.\footnote{We notice that some concurrent works~\cite{budzianowski2019hello,olabiyi2019multi,zhang2019dialogpt} also adopt the strategy similar to ours after the submission.}\\

\begin{figure*}
    \centering
    \includegraphics[width=1.5\columnwidth]{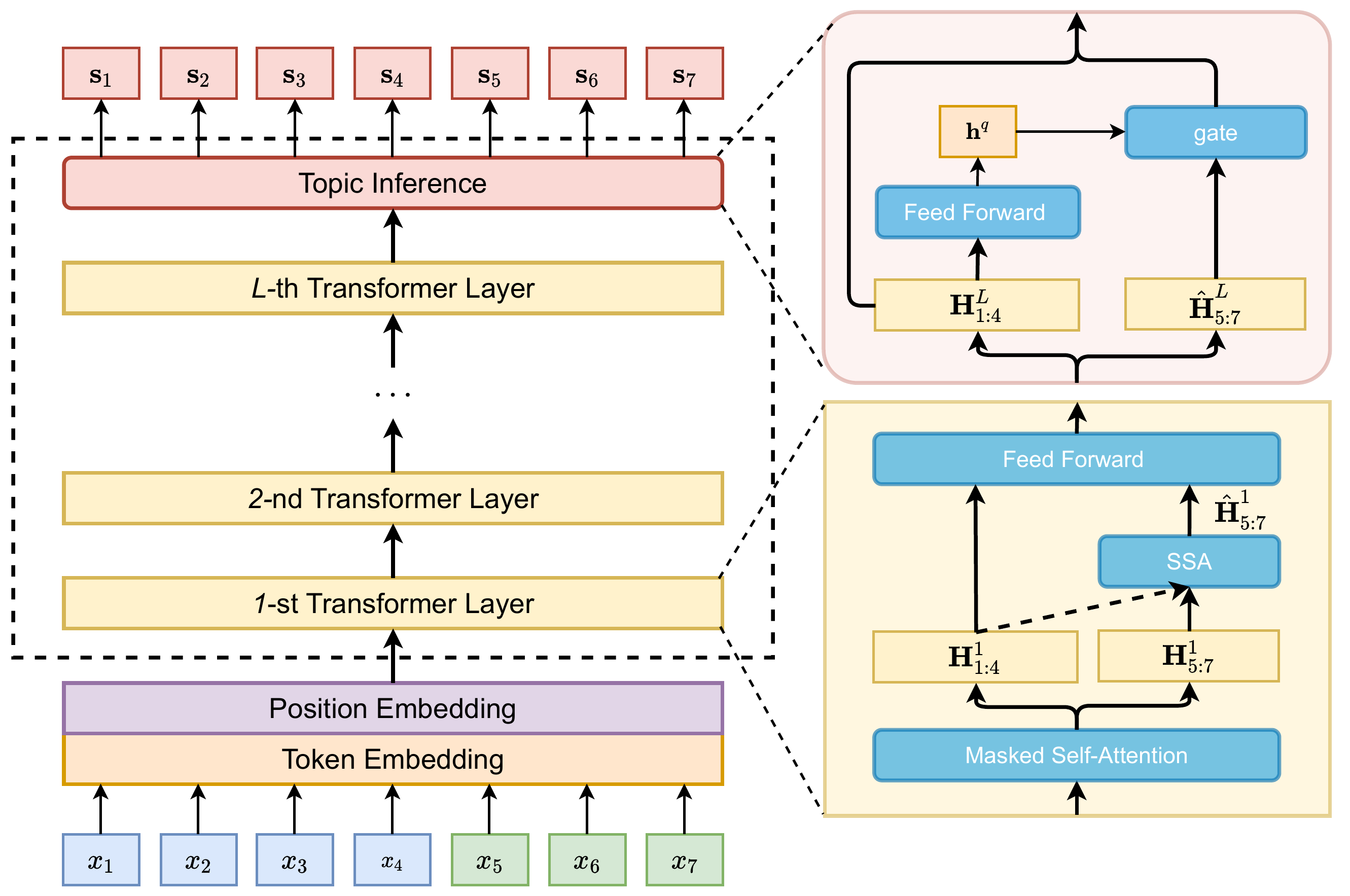}
    \caption{Overall architecture. The Topic Inference (TI) component on top of the transformer layers and the Supervised Source Attention (SSA) component inside the transformer layers are the proposed relevance-promoting components. Training losses are calculated on top of the obtained representation vectors $\mathbf{s}_t$'s.}
    \label{fig:architecture}
\end{figure*}

\section{Model}
\subsection{Overview}
In our language model formulation, each training query-response pair and the special tokens are concatenated as a single sequence $\mathrm{\bf x} = \{x_1,\cdots,x_{m},x_{m+1},\cdots,x_{n}\}$ of length $n$. $\mathrm{\bf x}_{1:m}$ corresponds to the query token sequence of length $m$ and $x_m$ is the special token [EOQ], denoting the end of query. $\mathrm{\bf x}_{m+1:n}$ corresponds to the response and $x_n$ is [EOS], the end symbol of the whole sequence. The training objective of our model is to maximize the unconditional likelihood $p(\mathrm{\bf x}_{1:n})$, similar to the existing language models~\cite{bengio2003neural,merity2018regularizing}. 

The architecture of our model is depicted in Fig~\ref{fig:architecture}, where $L$ decoder-only transformer layers~\cite{vaswani2017attention}\footnote{For the technical details of transformer, we recommend the reader to read the paper~\cite{vaswani2017attention}.} are involved. Different from the original transformer layer solely containing the self-attention component, the transformer layer in our model is further empowered with the proposed supervised source attention (SSA) component. The outputs of the $l$-th transformer layer are the contextualized token representations of size $\mathrm{dim}_h$, denoted as $\mathrm{\bf H}^{l} \in \mathbb{R}^{n \times \mathrm{dim}_h}$. When predicting the tokens, a Topic Inference (TI) component is introduced to provide the refined query representations encoding the topic information inferred from the reference.

\subsection{Language Model as Response Generator}
\label{subsection:lm_generator}
To achieve the goal of adapting language model for the STC task, we should carefully design a training strategy different from that in the \textsc{Seq2Seq} framework. Based on the observation that the human conversations can be regarded as a process of text continuation (i.e., giving the response/answer right after the query/question), we concatenate the query token sequence and the response token sequence into a single sequence and formulate the STC task as a contextual text continuation problem. One input example of our model is illustrated in Fig~\ref{fig:example_input}. The training goal of the model is to minimize the joint negative log likelihood over the whole sequence:
\begin{equation}
\begin{split}
    \mathcal{L}^{\mathrm{mle}} = -\log P(\mathrm{\bf x}_{1:n}) = -\sum^{n}_{t=1} \log P(\mathrm{\bf x}_t|\mathrm{\bf x}_{<t})
\end{split}
\label{eq:mle}
\end{equation}
Obviously, it is easy to bridge the gap between the task-specific training and the auto-regressive pre-training~\cite{peters-etal-2018-deep,radford2018improving,radford2019language} because the formulations of their objectives are almost the same. Another advantage of this language model formulation is that it takes the likelihood of query tokens into consideration, which is ignored in the existing works~\cite{shang-etal-2015-neural,xing2017topic}. Intuitively, the text generated by the language model is more fluent than those generated by \textsc{Seq2Seq} framework because the generator of the language model (the language model itself) is not only trained on the response sentence but also the query sentence. 

\subsection{Relevance Modeling Component}
The vanilla transformer decoder is equipped with self-attention~\cite{cheng-etal-2016-long,lin2017structured} and can theoretically capture the context of arbitrary length. Given the input $\mathrm{\bf H}^{l-1} \in \mathbb{R}^{n \times \mathrm{dim}_h}$, the contextualized representations $\mathrm{\bf h}^{l}_{t}$ ($l \in [1, L]$, $t \in [1,n]$) at the $t$-th time step is built as follows:
\begin{equation}
\begin{split}
    \mathrm{\bf h}^{l}_{t}, \pmb{\alpha}^{l}_t &= \textsc{Slf-Att} (\mathrm{\bf q}^{l-1}_{t}, \mathrm{\bf K}^{l-1}_{\leq t}, \mathrm{\bf V}^{l-1}_{\leq t}) \\
    \mathrm{\bf Q}^{l-1} &=  \mathrm{\bf H}^{l-1} \mathrm{\bf W}^{Q} \\
    \mathrm{\bf K}^{l-1}, \mathrm{\bf V}^{l-1} &= \mathrm{\bf H}^{l-1}\mathrm{\bf W}^{K}, \mathrm{\bf H}^{l-1} \mathrm{\bf W}^{V}
\end{split}
\end{equation}
where \textsc{Slf-Att} is the self-attention layer\footnote{The symbols for the feed-forward layer and residual connections are not shown.} and $\pmb{\alpha}^{l}_t \in \mathbb{R}^{t}$ is the calculated attention vector. $\mathrm{\bf Q}$, $\mathrm{\bf K}$, $\mathrm{\bf V} \in \mathbb{R}^{n \times \mathrm{dim}_h}$ respectively denote the query\footnote{Here, the ``query'' refers to a real-valued vector while the ``query'' in the STC task is a sentence.}, key and value in the self-attention layer. $\mathrm{\bf K}^{l-1}_{\leq t} = \{\mathrm{\bf k}^{l-1}_1,\cdots,\mathrm{\bf k}^{l-1}_t\}$ indicate the leftward elements and the same to $\mathrm{\bf V}^{l-1}_{\leq t}$. Despite its capability of learning global dependency, the transformer-based language model still has the tendency to overly rely on the short-term context and ignore the long-term context when predicting the next word, dubbed as \textit{\textbf{explanation away}} problem~\cite{holtzman2019curious}. This problem is catastrophic for the STC task because the query acts as the long-term context in our language model formulation and not involving the query information is prone to generating the content irrelevant to the query. Therefore, explicitly modeling the relevance and emphasizing the importance of the query are essential. In this paper, we propose two components, namely, \textbf{S}upervised \textbf{S}ource \textbf{A}ttention (SSA) and \textbf{T}opic \textbf{I}nference (TI), to handle the \textit{\textbf{explanation away}} problem. 

\paragraph{\textbf{Supervised Source Attention}} In the existing \textsc{Seq2Seq}-based frameworks, incorporating the query/source information is achieved by applying encoder-decoder attention solely on the encoder hidden representations. Similarly, attending only on the long-term context of language model is presumably beneficial for improving the relevance. Therefore, we propose to introduce another source attention layer on top of the self-attention layer. The computational formula of the $t'$-th ($t' \geq m$) query-enhanced hidden representation $\hat{\mathrm{\bf h}}^{l}_{t'}$ is below: 
\begin{equation}
\begin{split}
    \hat{\mathrm{\bf h}}^{l}_{t'}, \pmb{\beta}^l_{t'} &= \textsc{Src-Att}(\hat{\mathrm{\bf q}}^{l}_{t'}, \hat{\mathrm{\bf K}}^{l}, \hat{\mathrm{\bf V}}^{l}) \\
    \hat{\mathrm{\bf Q}}^{l} &= \mathrm{\bf H}^{l}\mathrm{\bf W}^{Q}  \\
    \hat{\mathrm{\bf K}}^{l}, \hat{\mathrm{\bf V}}^{l} &= \mathrm{\bf H}^{l}_{1:m}\mathrm{\bf W}^{K}, \mathrm{\bf H}^{l}_{1:m}\mathrm{\bf W}^{V}
\end{split}
\end{equation}
$\textsc{Src-Att}$ refers to our source attention layer on top of the self-attention layer. $\pmb{\beta}^{l}_{t'} \in \mathbb{R}^{m}$ is the attention scores for the corresponding hidden representations of the query tokens. $\mathrm{\bf H}^{l}$ is the output of \textsc{Slf-Att} layer and $\hat{\mathrm{\bf Q}}^{l} \in \mathbb{R}^{n \times \mathrm{dim}_h}$, $\hat{\mathrm{\bf K}}^{l}$, $\hat{\mathrm{\bf V}}^{l} \in \mathbb{R}^{m \times \mathrm{dim}_h}$ are the corresponding query, key, value in the source attention. Note that we only additionally apply source attention when the current token is not query token, i.e., $t' \geq m$, and do nothing in the preceding steps. Learning word alignment from data is possible but may be inaccurate without any supervision or external knowledge~\cite{liu-etal-2016-neural,mi-etal-2016-supervised}, therefore, we employ the keywords as the knowledge and enforce the source attention component to be concentrated on the important query tokens. First of all, we perform \textbf{max-over-time} pooling over the attention vectors $\pmb{\beta}^l_{t'}\in \mathbb{R}^m$ ($t' \in [m+1, n]$) and induce the vector $\hat{\mathrm{\bf y}}^{\mathrm{src}} \in \mathbb{R}^m$ reflecting the salience scores of the query/source tokens:
\begin{equation}
    \hat{\mathrm{\bf y}}^{\mathrm{src}}_{i} = \max\{\pmb{\beta}^L_{m+1,i},\cdots,\pmb{\beta}^L_{n,i}\}, i \in [1, m]
\end{equation}
Then, given the query keyword indicator vector $\mathrm{\bf y}^{\mathrm{src}} \in \{0,1\}^m$, we introduce additional source attention loss $\mathcal{L}^{src}$ into Eq~(\ref{eq:mle}):
\begin{equation}
    \mathcal{L}^{\mathrm{src}} = \frac{1}{m} ||\hat{\mathrm{\bf y}}^{\mathrm{src}}_{i}-\mathrm{\bf y}^{\mathrm{src}}||^2_2
    \label{eq:src_loss}
\end{equation}
Ideally, the generation process will rely on more important query tokens if the salience score $\hat{\mathrm{\bf y}}^{\mathrm{src}}$ is more close to the keyword vector $\mathrm{\bf y}^{\mathrm{src}}$.

\paragraph{\textbf{Topic Inference}} 
The SSA component attempts to improve the relevance by highlighting the importance of the important query tokens/words in the attention process. However, the range of the words topically related to the query is far more than that of the keywords explicitly mentioned in the query. Considering the query ``\textit{what is your favorite fruit?}'' and two valid responses ``\textit{I like the watermelon very much}'' and ``\textit{My favorite fruit is pineapple}'', ``fruit'' should be emphasized during the generation but the words used to discuss fruit such as ``watermelon'' and ``pineapple'' are also very meaningful for building a response. Inspired by this, we collect the multiple references of each query in the training set and gather all of the keywords extracted from such responses\footnote{\cite{xing2017topic} extend the keyword set using external corpus. Here, we focus on improving the relevance rather than enriching the topical words in the response, thus, we only utilize the training data to explore more keywords.}. To exploit the latent topic information, we introduce Topic Inference (KI) component to estimate the global topical word distribution based on the query representation $\mathrm{\bf h}^q$ as follows:
\begin{equation}
    \begin{split}
        \mathrm{\bf h}^q = f(\mathrm{\bf x}_{1:m}) , \quad P(z|\mathrm{\bf x}_{1:m}) = \text{Softmax}(\mathrm{\bf W}^{o} \mathrm{\bf h}^q)
    \end{split}
\label{eq:softmax}
\end{equation}
where $f:\mathbb{R}^{m} \to \mathbb{R}^{\mathrm{dim}_h}$ denotes the function mapping the input query tokens to a low-dimensional query representation. Specifically, we feed the last query hidden representation in the transformer, namely, $\mathrm{\bf h}^L_m$, into a linear layer with \texttt{tanh} activation and regard the output as the query representation $\mathrm{\bf h}^q$ for simplifying the modeling part. To encode the topic information into the query representation, we employ the global keyword indicator vector $\mathrm{\bf y}^{\mathrm{kwd}} \in \{0,1\}^{|\mathcal{V}|}$ as supervision signals and enforce the components corresponding to keywords/important tokens in the query-based global topic distribution to be up-weighted. The computational formula is as follows:
\begin{equation}
    \mathcal{L}^{\mathrm{kwd}} = -\frac{1}{|\mathcal{V}|} \sum^{|\mathcal{V}|}_{i=1} \mathrm{\bf y}^{\mathrm{kwd}}_{i} \cdot \log P_i(z|\mathrm{\bf x}_{1:m})
    \label{eq:keyword}
\end{equation}
where the subscript $i$ denotes the $i$-th component of a vector and $|\mathcal{V}|$ is the vocabulary size. Note that we attempt to replace the Softmax in Eq~\ref{eq:softmax} with the component-wise Sigmoid, typically used in multi-label classification problem, but the empirical results become worse. Thus, we keep the Softmax probability function unchanged in the experiment. Similar to Eq~\ref{eq:src_loss}, the $\mathcal{L}^{\mathrm{kwd}}$ will  be added in the training loss. 

Different from~\cite{yao-etal-2017-towards} and~\cite{gao2019generating} regarding the concrete topic/keyword as the trigger of generation, we introduce the query representation encoding the global topic information as the supplementation for each token-level representation to encourage the generation of the relevant topical words. The representation vector $\mathrm{\bf s}_t$ for predicting the output is calculated below: 
\begin{equation}
    \begin{split}
        \mathrm{\bf s}_t &= \left\{\;
            \begin{array}{cl}
              (1-g_t) * \mathrm{\bf h}^L_t + g_t * \mathrm{\bf h}^q & \text{, if } t > m  \\ 
              \mathrm{\bf h}^L_t & \text{, Otherwise}  \\
            \end{array}
            \right. 
            \\
        g_t &= \sigma(\mathrm{\bf W}^g \mathrm{\bf h}^q + \mathrm{\bf W}^l \mathrm{\bf h}^L_t + \mathrm{\bf b}), 
    \end{split}
\end{equation}
where $ g_t \in \mathbb{R}^{\mathrm{dim}_h}$ is the gate value and $\mathrm{\bf W}^g, \mathrm{\bf W}^l \in \mathbb{R}^{\mathrm{dim}_h \times \mathrm{dim}_h}$ are parameter matrices in the TI component. 

\subsection{Model Training}
The proposed SSA component and the TI component are jointly trained with the transformer-based language model. Based on Eq~\ref{eq:mle}, Eq~\ref{eq:src_loss} and Eq~\ref{eq:keyword}, the overall training objective $\mathcal{L}(\theta)$ of the proposed model is as follow: 
\begin{equation}
    \begin{split}
        & \mathcal{L}(\theta) = \frac{1}{|\mathbb{D}|} \sum_{(\mathrm{\bf x}, \mathrm{\bf y}^{\mathrm{src}}, \mathrm{\bf y}^{\mathrm{kwd}}) \in \mathbb{D}} \mathcal{L}(\mathrm{\bf x}, \mathrm{\bf y}^{\mathrm{src}}, \mathrm{\bf y}^{\mathrm{kwd}}) \\
        & \mathcal{L}(\mathrm{\bf x}, \mathrm{\bf y}^{\mathrm{src}}, \mathrm{\bf y}^{\mathrm{kwd}}) = \mathcal{L}^{\mathrm{mle}} + \gamma_1 \mathcal{L}^{\mathrm{src}} + \gamma_2 \mathcal{L}^{\mathrm{kwd}}
    \end{split}
\end{equation}
Here, $\gamma_1$ and $\gamma_2$ are the coefficients controlling the proportion of $\mathcal{L}^{\mathrm{src}}$ and $\mathcal{L}^{\mathrm{kwd}}$ involved in the training respectively.

\subsection{Decoding}
Due to the limited search space, it is difficult for the beam search or greedy search to find the interesting and diverse responses. Therefore, we do not adopt them but a ``\textit{\textbf{randomization-over-maximization}}'' strategy (also know as `top-\textit{k} sampling'') to perform the decoding, as done in~\cite{fan-etal-2018-hierarchical,radford2019language}. \cite{holtzman2019curious} and~\cite{ippolito-etal-2019-comparison} explore the usage of other advanced decoding strategies in the language generation task. Since our aim in this paper is not to compare the performances across the different decoding strategies, we consistently use the top-\textit{k} sampling.

\section{Experiment}
\subsection{Experiment Setup}
We utilize the benchmark STC dataset~\cite{liu-etal-2018-towards-less} to evaluate the effectiveness of the proposed relevance-promoting transformer language model. This dataset is built based on the real conversations from \texttt{Weibo}\footnote{https://www.weibo.com/} and contains about 7M high-quality query-response pairs. We split the dataset such that \#train:\#dev:\#test is 7,024,156:2,000:800. Training details are provided in the appendix.

To avoid word segmentation errors and out-of-vocabulary issue, the tokens in our model and the baseline models are Chinese characters and the vocabulary size is about 12,000.

\subsection{Evaluation Metrics}
We introduce the following metrics to evaluate the model's capability of generating relevant and diverse responses:

\noindent
{\bf Relevance Metrics} We employ \textbf{\textsc{Bleu-2}}, \textbf{\textsc{Bleu-3}} \& \textbf{\textsc{Bleu-4}}~\cite{papineni-etal-2002-bleu} to estimate the relevance of the generated responses. Moreover, we also design two more metrics, namely, \textbf{\textsc{Hit-q}} and \textbf{\textsc{Hit-r}} to calculate the hit rates of the topical words in the query and the response respectively. Firstly, we build a \textit{high-precision-low-recall} keyword set for each query/response sentence based on keyword extraction toolkit\footnote{https://github.com/fxsjy/jieba} and filter some noisy words based on additional hand-crafted rules. Then, we calculate the \textsc{Hit-Q}$_i$ and \textsc{Hit-R}$_i$ for the $i$-th predictions as follows:
\begin{equation}
    \textsc{Hit-q}_i = \frac{|\mathbb{K}^{r_i} \cap \mathbb{K}^{q_i}|}{|\mathbb{K}^{r_i}|},\textsc{Hit-r}_i = \frac{|\mathbb{K}^{r_i} \cap \mathbb{K}^{r^g_i}|}{|\mathbb{K}^{r_i}|}
\end{equation}
where $\mathbb{K}^{q_i}$, $\mathbb{K}^{r_i}$ and $\mathbb{K}^{r^g_i}$ respectively denote the topical word set for the $i$-th query, predicted response and gold standard response. Then we obtain the \textsc{Hit-q} and \textsc{Hit-r} by performing the corpus-level average:
\begin{equation}
    \textsc{Hit-q} = \frac{1}{N}\sum^N_i\textsc{Hit-q}_i, \textsc{Hit-r} = \frac{1}{N}\sum^N_i\textsc{Hit-r}_i
\end{equation}

\noindent
{\bf Diversity Metrics}  Following~\cite{li-etal-2016-diversity}, we employ \textbf{\textsc{Dist-1}} and \textbf{\textsc{Dist-2}} to calculate the ratios of the distinct uni-grams and bi-grams in the generated responses.

\noindent
{\bf Human Evaluations}  We also conduct human evaluations. Specifically, we randomly sampled 100 queries and recruit five helpers to judge \textit{Relevance} (4-scale rating, 0-3), \textit{Fluency} (3-scale rating, 0-2) and \textit{Acceptance} (0 or 1) of the generated responses from our model and the baselines. Details of the rating criteria are stated in the appendix.

\subsection{Comparison Models}
\begin{itemize}[leftmargin=*]
    \setlength\itemsep{0pt}
    \item \textbf{LSTM-LM}~\cite{mei2017coherent}: LSTM-based auto-regressive language model armed with incremental self-attention. We train LSTM-LM using the same strategy mentioned in this paper.
    \item \textbf{LSTM-S2S}: Attention-based LSTM Sequence-to-Sequence model. 
    \item \textbf{TFM-S2S}: Transformer Sequence-to-Sequence model where the network components are identical to those in~\cite{vaswani2017attention}. 
    \item \textbf{TFM-LM}: Transformer-based auto-regressive language model. We train TFM-LM using the same strategy mentioned in this paper. 
    \item \textbf{MMI}~\cite{li-etal-2016-diversity}: LSTM-S2S with Maximum Mutual Information objective in decoding. In this paper, we set the number of responses for re-ranking as 50.
    \item \textbf{CVAE}~\cite{zhao-etal-2017-learning}\footnote{https://github.com/snakeztc/NeuralDialog-CVAE}: Conditional Variational Auto-Encoder for response generation. We replace the dialogue acts used in the original model with the keywords extracted from the references.
    \item \textbf{MMPMS}~\cite{chen2019generating}: The model with the  state-of-the-art performance on the STC task. We re-run the officially released code\footnote{https://github.com/PaddlePaddle/models} to obtain the results on our dataset.
\end{itemize}

\begin{table*}[]
    \centering
    \resizebox{1.7\columnwidth}{!}{
    \begin{tabular}{l|ccccc|cc}
    \Xhline{3\arrayrulewidth}
     \multirow{2}{*}{\textbf{Model}} & \multicolumn{5}{c|}{\textbf{Relevance}} & \multicolumn{2}{c}{\textbf{Diversity}} \\ \cline{2-8}
    & \textsc{Bleu-2} & \textsc{Bleu-3} & \textsc{Bleu-4} & \textsc{Hit-Q} & \textsc{Hit-R} & \textsc{Dist-1} & \textsc{Dist-2} \\ \hline \hline
    LSTM-LM & 3.8 & 0.9 & 0.3 & 0.084 & 0.066 & 0.028 & 0.094 \\
    LSTM-S2S & 5.6 & 2.8 & 1.8 & 0.293 & 0.145 & 0.039 & 0.137 \\
    TFM-LM & 6.9 & 3.2 & 2.1 & 0.295 & 0.144 & 0.058 & 0.259 \\
    TFM-S2S & 7.3 & 3.5 & 2.3 & 0.369 & 0.172 & 0.078 & 0.290 \\
    MMI & 7.9 & 2.5 & 1.0 & 0.197 & 0.145 & 0.093 & 0.349 \\
    CVAE & 5.8 & 1.5 & 0.4 & 0.211 & 0.135 & 0.060 & 0.211 \\
    MMPMS & 6.7 & 3.0 & 1.8 & 0.151 & 0.102 & 0.057 & 0.220 \\ \hline \hline
    OURS-\texttt{tk} w/o SSA \& TI & 4.9 & 1.0 & 0.3 & 0.119 & 0.076 & 0.086 & 0.441 \\  
    OURS-\texttt{tk} w/o SSA & 5.5 & 2.1 & 1.5 & 0.150 & 0.146 & 0.102 & 0.521\\   
    OURS-\texttt{tk} w/o TI & 5.1 & 2.1 & 1.4 & 0.171 & 0.132 & 0.090 & 0.445 \\  
     \hline \hline
    OURS-\texttt{bm} & \textbf{10.3} & \textbf{5.3} & \textbf{3.4} & \textbf{0.510} & \textbf{0.193} & 0.102 & 0.398 \\ 
    OURS-\texttt{tk} & 6.0 & 3.6 & 2.5 & 0.191 & 0.152 & \textbf{0.107} & \textbf{0.544} \\
    \Xhline{3\arrayrulewidth}
    \end{tabular}}
    \caption{Experimental results on the automatic metrics. The best results are in \textbf{bold}.}
    \label{tab:automatic}
\end{table*}

\begin{table}[t]
    \centering
    \resizebox{0.9\columnwidth}{!}{
    \begin{tabular}{l|ccc}
    \Xhline{3\arrayrulewidth}
        \multirow{2}{*}{\textbf{Model}} & \multicolumn{3}{c}{Evaluation Metrics}  \\ \cline{2-4}
         & \textit{Relevance} & \textit{Fluency} & \textit{Acceptance} \\ \hline \hline
         LSTM-LM & 1.206 & 1.297 & 0.26 \\
         LSTM-S2S & 1.386 & 1.285 & 0.37 \\
         TFM-LM & 1.412 & 1.328 & 0.39 \\ 
         TFM-S2S & 1.475 & 1.306 & 0.43 \\ 
         MMI & 1.432 & 1.301 & 0.34 \\
         CVAE & 1.316 & 1.274 & 0.33 \\
         MMPMS & 1.528 & 1.396 & 0.42 \\ \hline \hline
         OURS-\texttt{tk} w/o SSA \& TI & 1.273 & 1.368 & 0.28\\ 
         OURS-\texttt{tk} w/o SSA & 1.485 & \textbf{1.407} & 0.39 \\
         OURS-\texttt{tk} w/o TI & 1.503 & 1.303 & 0.36 \\ \hline \hline
         OURS-\texttt{bm} & 1.515 & 1.359 & 0.38 \\
         OURS-\texttt{tk} & \textbf{1.606} & 1.346 & \textbf{0.44} \\ 
    \Xhline{3\arrayrulewidth}     
    \end{tabular}}
    \caption{Human evaluation results with the best ones in \textbf{bold}.}
    \label{tab:human}
\end{table}

\subsection{Main Results}
Table~\ref{tab:automatic} and~\ref{tab:human} list the automatic evaluation results and the human evaluation results respectively. In terms of \textsc{Bleu}, the proposed model with beam search decoding, namely, OURS-\texttt{bm}, consistently achieve the best scores. Besides, OURS-\texttt{bm} outperforms all compared models on the keyword-overlapping-based \textsc{Hit} metrics, suggesting that our model, armed with Supervised Source Attention component (SSA) and Topic Inference (TI) component, is beneficial for the generation of informative topical words related to the query. Surprisingly, OURS-\texttt{bm} also obtains better \textsc{Dist} metrics than the baseline models. After replacing the beam search with top-\textit{k} sampling, our model (OURS-\texttt{tk}) is further enhanced in diversity modeling, reaching 0.107 and 0.544 on  \textsc{Dist-1} and \textsc{Dist-2} respectively. 

Regarding the more reliable human evaluations, both of OURS-\texttt{bm} and OURS-\texttt{tk} are the top-ranked models. Specifically, despite its unsatisfactory results on the automatic \textsc{Bleu} and \textsc{Hit} metrics, OURS-\texttt{tk} performs the best on the manually annotated \textit{Relevance} metric with 5\% improvement over the current state-of-the-art MMPMS model. Instead, OURS-\texttt{bm}, the best model on the automatic relevance metrics, still yields competitive results on the \textit{Relevance}. It is reasonable because some words not appearing in the query/references, especially those not being frequently used, are still related to the discussed topic in the conversations. At the same time, such inconsistency between automatic and human evaluations demonstrates the effectiveness of top-\textit{k} sampling, a \textit{\textbf{randomization-over-maximization}} decoding strategy, in discovering infrequent but meaningful patterns for the STC task.  

We now turn to discuss the performance of the other compared methods. Inheriting the powerful modeling capability of Transformer, TFM-S2S obtains the best automatic relevance scores as well as the second best \textit{Relevance} among the baselines. TFM-LM, another Transformer-based baseline following the language model formulation in our paper, performs not as good as TFM-S2S on all of the metrics except \textit{Fluency}, verifying the postulation that the \textbf{\textit{explanation away}} issue of language model has the tendency to produce fluent but topically irrelevant responses. Despite of this, the TFM-LM outperforms LSTM-LM and LSTM-S2S, proving the superiority of Transformer to LSTM in response generation. Owing to the re-ranking mechanism, the MMI model is the strongest baseline on diversity modeling but OURS-\texttt{bm}/OURS-\texttt{tk} still achieves approximately 14\%/55\% improvement on \textsc{Dist-2}.

\subsection{Ablation Study}
In order to track the source of the performance gains, we also conduct the ablation study on the OURS-\texttt{tk}. The corresponding automatic and human evaluation results are shown in the second group of Table~\ref{tab:automatic} and Table~\ref{tab:human}. As expected, the model without relevance-promoting design, i.e., OURS-\texttt{tk} w/o SSA \& TI, is the worst one on the relevance metrics. OURS-\texttt{k} w/o SSA and OURS-\texttt{tk} w/o TI, the variants incorporating either TI or SSA for relevance modeling, boost the \textit{Relevance} score by $\sim$17\% and $\sim$18\% respectively. Although they are comparable on the relevance metrics but the former achieves higher diversity scores (\textsc{Dist-2}: 0.521 v.s. 0.441). We attribute this phenomenon to the TI component, which exploits the usage of more related topical words mentioned in the multiple references. With the help of both SSA component and TI component, OURS-\texttt{tk} becomes the best model on \textit{Relevance} and \textsc{Dist} metrics, demonstrating the necessity of the relevance modeling for the transformer language model. Another interesting finding is that the SSA component decreases the \textit{Fluency} score (see the results of OURS-\texttt{tk} w/o TI), which indicates that fighting against \textit{\textbf{explanation-away}} issue by incorporating additional query context may be coupled with corrupting the language model.  

\begin{figure*}
    \centering
    \includegraphics[width=1.97\columnwidth]{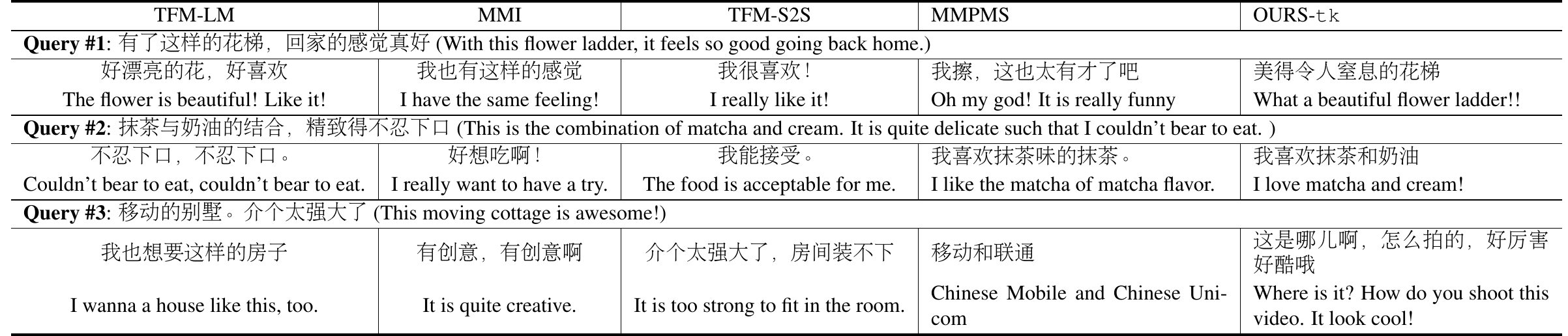}
    \caption{Examples of response generation. We translate Chinese samples to English.}
    \label{fig:case_analysis}
\end{figure*}
\subsection{Case Study}
Figure~\ref{fig:case_analysis} shows example responses generated by our model and the most competitive baseline models. OURS-\texttt{tk}, which explicitly incorporates the query context and exploits the tokens potentially related to the query, always produces meaningful and informative responses. Taking the Query \#1 \& \#2 as examples, the generated responses accurately respond to the query because they mention ``flower ladder''/``matcha'' and ``cream'', which are exactly the topics discussed in the conversations. The response for the Query \#3 can easily engage user in the conversation and thus it is also a meaningful prediction. The outputs of TFM-LM are generally fluent. However, due to the \textit{\textbf{explanation away}} issue, TFM-LM tends to generate the irrelevant response (Case \#1) or response with phrase repetition (Case \#2). Under the sequence-to-sequence formulation, TFM-S2S obtains the responses moderately related to the corresponding queries although the third output, directly copying part of the source text (i.e., query), is still unsatisfactory. MMPMS and MMI, the models aiming for promoting diversity, have chances to yield irrelevant responses. 

\subsection{Further Discussions on Top-\textit{k} Sampling}
We further investigate the impact of top-\textit{k} sampling on the STC models. Firstly, we conduct additional automatic and human evaluations on the baseline models with results shown in Table~\ref{tab:further_comparison}. As can be seen, the top-\textit{k} sampling consistently improves the \textsc{Dist-2} score by a large margin on all models but the \textit{Relevance} scores of LSTM-S2S, TFM-LM and TFM-S2S decrease after top-\textit{k} sampling is applied. The variation trends of \textit{Fluency} across the evaluated models are also inconsistent. These observations suggest that top-\textit{k} sampling is simple yet effective to achieve diverse response generation but it should be carefully utilized in the model because of its uncertainty on relevance and fluency. 

\begin{figure}
    \centering
    \includegraphics[width=1\columnwidth]{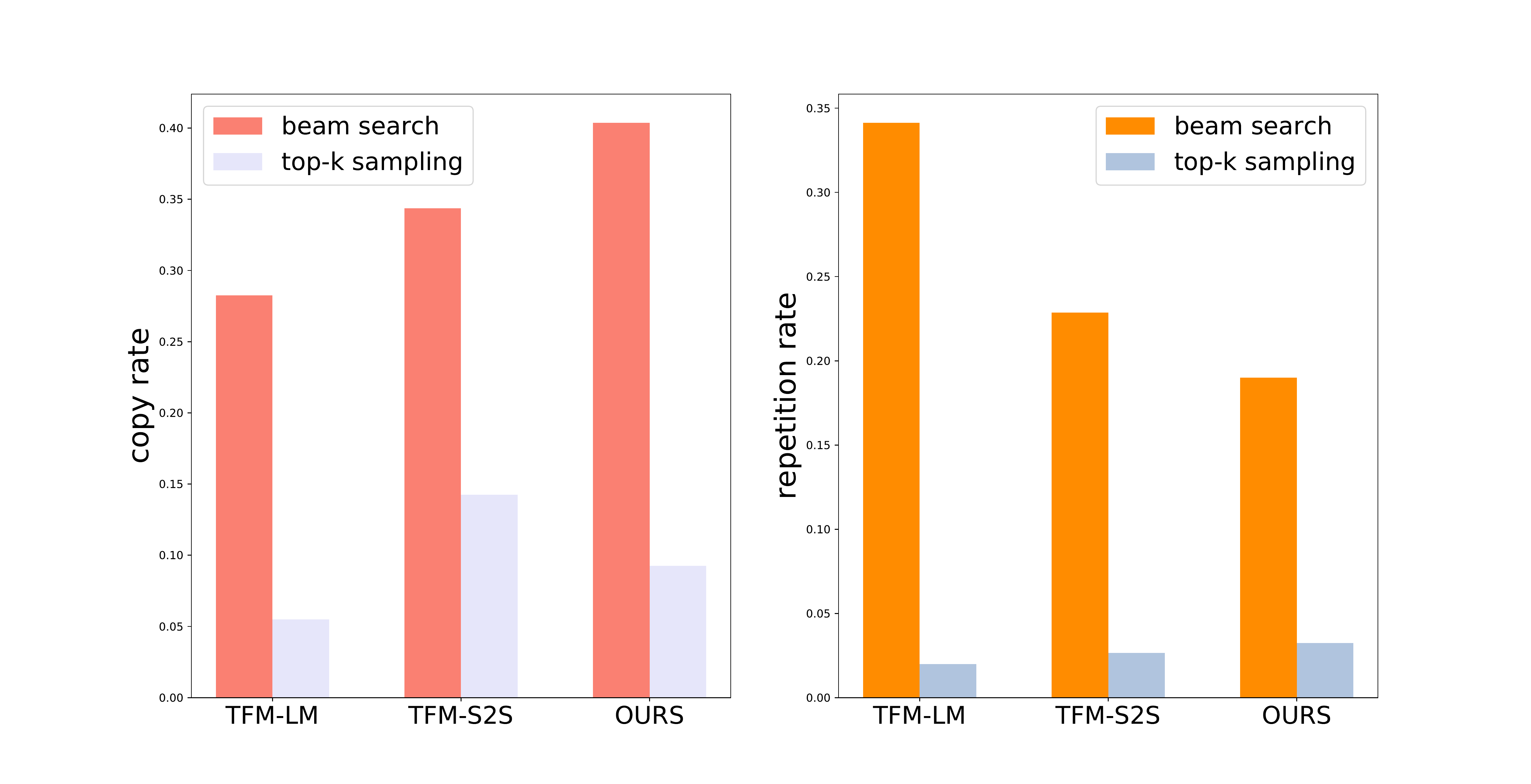}
    \caption{Comparison results on beam search and top-\textit{k} sampling. Specifically, if the length of the longest common sub-string between response and query is larger than 4, then the response is regarded as a ``copy'' of query. If a response contains the word/phrase loop over 3 times, it is regarded as a response with repetition.}
    \label{fig:tk_analysis}
\end{figure}

As discussed in Case Study, the transformer-based models adopting beam search have the tendency to generate the responses with repetition and those directly copying the query. 
We here investigate whether top-\textit{k} sampling can help solve these issues. Figure~\ref{fig:tk_analysis} depicts the ratios of responses in the test set falling into the phrase repetition and query copy. The top-\textit{k} sampling greatly reduces the query copy rate (about 72\% on average) and almost eliminates the phrase repetition phenomenon in the Transformer-based models. However, note that  Table~\ref{tab:further_comparison} shows both TFM-LM and TFM-S2S perform worse on \textit{Relevance} after using top-\textit{k} sampling. We consider these results are consistent with human perception because enriching the morphology via sampling-based decoding strategy will inevitably introduce irrelevant information, leading to the degradation of relevance score. It is noticeable that the proposed model (i.e., OURS) is not affected on relevance modeling due to its capability of filtering some topically irrelevant candidates for the sampling process.

\begin{table}[h]
    \centering
    \resizebox{1\columnwidth}{!}{
    \begin{tabular}{l|lll}
    \hline 
        Models & \textit{Relevance} ($\Delta$) & \textit{Fluency} ($\Delta$) & \textsc{Dist-2} ($\Delta$) \\ \hline
        LSTM-LM-\texttt{tk} & 1.111 (-0.09) & 1.270 (-0.03) & 0.383 (+0.29)\\  
        LSTM-S2S-\texttt{tk} & 1.439 (+0.05) & 1.265 (-0.20) & 0.490 (+0.35) \\
        TFM-LM-\texttt{tk} & 1.273 (-0.14) & \textbf{1.368} (+0.04) & 0.441 (+0.18) \\ 
        TFM-S2S-\texttt{tk} & 1.270 (-0.15) & 1.321 (+0.15) & 0.507 (+0.22) \\
        OURS-\texttt{tk} & \textbf{1.606} (+0.10) & 1.346 (-0.13) & \textbf{0.544} (+0.20) \\
    \hline 
    \end{tabular}}
    \caption{Experimental results on the models adopting top-\textit{k} sampling. $\Delta$ refers to the improvement over the original model adopting beam search. The best results are in \textbf{bold}.}
    \label{tab:further_comparison}
\end{table}

\section{Related Work}
{\bf Short Text Conversation} 
Short Text Conversation (STC) is usually formulated as a conditional text generation task~\cite{shang-etal-2015-neural,serban2016building}. The sequence-to-sequence (\textsc{Seq2Seq}) encoder-decoder framework~\cite{cho-etal-2014-learning,sutskever2014sequence,bahdanau2015neural} and its variants have been studied extensively for solving this task. \citeauthor{li-etal-2016-diversity}~\citeyear{li-etal-2016-diversity} introduce diversity-promoting decoding strategies into the \textsc{Seq2Seq} model. Some \cite{mou-etal-2016-sequence,xing2017topic,yao-etal-2017-towards,zhou2017mechanism,gao2019generating} attempt to guide the \textsc{Seq2Seq} model to generate keyword/topic-aware responses while others \cite{wu2019response,cai-etal-2019-skeleton,cai-etal-2019-retrieval} try to control the response generation with additional retrieved data. 
The advanced techniques such as RL, GAN and VAE are also considered for improving conversational experience~\cite{li-etal-2016-deep,xu-etal-2017-neural,du-etal-2018-variational,gao-etal-2019-discrete}.

\noindent
{\bf Transformer-based Language Model}
Deep transformer-based architecture~\cite{vaswani2017attention} has led to significant performance gains on the language modeling task~\cite{al2019character,dai-etal-2019-transformer,radford2019language}, compared to the existing CNN/RNN-based architectures~\cite{dauphin2017language,merity2018regularizing,melis2018state}. Meanwhile, GPT-2~\cite{radford2019language} and \textsc{UniLM}~\cite{dong2019unified} are the pioneer works adapting the transformer language model for the conditional text generation tasks.

\section{Conclusion}
In this paper, we present a language model based solution instead of traditional \textsc{Seq2Seq} paradigm for handling Short-Text Conversation (STC). We firstly tailor-make a training strategy to adapt the language model for the STC task. Then, we propose a relevance-promoting transformer language model to distill the relevance clues from the query as well as  the topics inferred from the references, and incorporate them into the generation. Moreover, we explore the usage of top-\textit{k} sampling for the STC task to further improve the response diversity.  Experimental results on a large-scale STC dataset validate that our model is superior to the compared models on both relevance and diversity from automatic and human evaluations.

\bibliographystyle{aaai}
\small
\bibliography{aaai20}

\section{Appendices}
\subsection{Training Details}
Our model consists of 6 decoder-only transformer layers with masked self-attention (i.e., $L$=6), where the hidden size $\mathrm{dim}_h$, number of heads and feed-forward size are 512, 8, 1024 respectively. The weights $\gamma_1, \gamma_2$ for $\mathcal{L}^{\mathrm{src}}$ and $\mathcal{L}^{\mathrm{kwd}}$ are set as 1.0 and 0.2. We do not introduce the pre-trained word/character embeddings but randomly initialize the parameters of the token embedding layer. We employ Adam~\cite{kingma2015adam} as optimizer and the initial learning rate is 1e-4. We apply linear warm-up at the first 10,000 training steps. The batch size is 32 and we train the model up to 20 epoch. We evaluate the model every 30,000 steps and select the model performs best on the validation set for producing the final results.

\subsection{Human Evaluations}
Apart from automatic evaluations, we also conduct human evaluations. Specifically, we randomly sampled 100 queries and recruit five helpers to judge \textit{Relevance}, \textit{Fluency} and \textit{Acceptance} of the generated responses from our model and the baselines. The rating criteria, identical to those in~\cite{liu-etal-2018-towards-less}, are as follows:

\indent $\bullet$ \textit{Relevance}: \textbf{+3}: relevant as well as interesting; \textbf{+2}: relevant, including the generic responses; \textbf{+1}: relevant at a distant level; \textbf{0}: not relevant at all. \\
\indent $\bullet$ \textit{Fluency}: \textbf{+2}: fluent; \textbf{+1}: readable but with some grammar mistakes; \textbf{0}: unreadable. \\
\indent $\bullet$ \textit{Acceptance}: the ratio of acceptable responses. Specifically, acceptable response refers to the response with \textit{Relevance} $\geq 2$ and \textit{Fluency} $\geq 1$. 

\subsection{Obtaining Informative Query Words}
Building the supervision signals $\mathrm{\bf y}^{\mathrm{src}}$ in Eq~\ref{eq:src_loss} is based on the informative words of each query. The basic idea is that a query word having strong semantic relation with the corresponding response should be regarded as an informative word. The procedure is as follows:
\begin{enumerate}
    \item Use keyword extractor\footnote{Here, we use jieba keyword extraction toolkit available at https://github.com/fxsjy/jieba.} to obtain the keywords for each response in the training set.
    \item Define the semantic relation score between a query word and the response as the maximal point-wise mutual information (PMI) between a query word and the response keywords.
    \item Select the top-ranking query words in terms of the calculated semantic relation scores as the informative words.
\end{enumerate}

\subsection{Obtaining Response Keywords}
The proposed Topic Inference (TI) component aims to refine the query representation with the knowledge inferred from response keywords. First of all, we employ jieba keyword extraction toolkit to collect the response keywords. Since one query may correspond to multiple references (i.e., \textbf{one-to-many} phenomenon), we aggregate the keyword sets for multiple responses corresponding to the same query. Then, we randomly sample \textbf{80\%} keywords in the aggregated set and regard them as the relevant response keywords $\mathrm{\bf y}^{\mathrm{kwd}}$ (in Eq~\ref{eq:keyword}) associated with each training instance.  

\subsection{Obtaining Keywords for Evaluation}
As mentioned in the Experiment part, calculating the \textsc{Hit-Q} and \textsc{Hit-R} metrics need to build a \textit{high-precision-low-recall} keyword set for each query/response sentence. We firstly employ jieba keyword extraction toolkit to obtain an initial keyword set for each query/response. Then, we design the following rules to guarantee the precision of the obtained query/response keywords:

\indent $\bullet$ Remove the stop words in the initial keyword set.\\
\indent $\bullet$ Filter the keyword if the Part-of-Speech tag of this keyword does not belong to \{\textit{N}, \textit{NS}, \textit{VN}, \textit{V}, \textit{F}\}.

\subsection{Additional Details of Experiment}
For the automatic evaluation results in Table~\ref{tab:automatic}, \textbf{\textsc{Bleu}} and \textbf{\textsc{Dist}} are character-level metrics while \textbf{\textsc{Hit}} scores are calculated using the word-based overlapping statistics. 

\end{document}